\documentclass[10pt, review]{elsarticle}

\usepackage{xcolor}
\usepackage{amsfonts}
\usepackage{graphicx}
\usepackage{amsmath}
\usepackage{amssymb}
\usepackage{multirow}
\usepackage{blindtext}
\usepackage{wrapfig,booktabs}
\usepackage{mathtools}
\usepackage{comment}
\usepackage{lipsum}
\usepackage{enumitem}
\usepackage{hyperref}
\usepackage{adjustbox}
\usepackage{subcaption}
\usepackage{algpseudocode}
\usepackage{algorithm}

\newcommand{\rpm}{\raisebox{.2ex}{$\scriptstyle\pm$}}

\usepackage{mfirstuc} 
\newcommand{\addReviewer}[2]{
	\expandafter\newcommand\csname #1\endcsname[1]{{\bf \color{#2} \capitalisewords{#1}:\,##1}}
	\expandafter\newcommand\csname #1cor\endcsname[2]{{\color{#2} \capitalisewords{#1}:\,\st{##1}{\bf ##2}}}
	\expandafter\newcommand\csname #1color\endcsname{#2}
}

\newif\ifproofread

\newcommand{\changemarker}[1]{%
\ifproofread
\textcolor{blue}{#1}%
\else
#1%
\fi
}

\newcommand{\changemarkertable}[1]{%
\ifproofread
{\color{blue}#1}%
\else
#1%
\fi
}

\definecolor{asparagus}{rgb}{0.53, 0.66, 0.42}
\definecolor{alizarin}{rgb}{0.82, 0.1, 0.26}
\addReviewer{jary}{asparagus}
\addReviewer{simone}{alizarin}

\newcommand{\nome}{Structured Ensemble}

\begin{document}
\proofreadfalse

\begin{frontmatter}
\title{{\nome}s: an Approach to Reduce the Memory Footprint of Ensemble Methods}

\author[sapienza]{Jary Pomponi \corref{cor1}}
\cortext[cor1]{Corresponding author: jary.pomponi@uniroma1.it}

\author[sapienza]{Simone Scardapane}

\author[sapienza]{Aurelio Uncini}

\address[sapienza]{Department of Information Engineering, Electronics and Telecommunications (DIET), Sapienza University of Rome, Italy}

		\begin{abstract}
			In this paper, we propose a novel ensembling technique for deep neural networks, which is able to drastically reduce the required memory compared to alternative approaches. In particular, we propose to extract multiple sub-networks from a single, untrained neural network by solving an end-to-end optimization task combining differentiable scaling over the original architecture, with multiple regularization terms favouring the diversity of the ensemble. Since our proposal aims to detect and extract sub-structures, 
			we call it \textbf{\nome{}}. On a large experimental evaluation, we show that our method can achieve higher or comparable accuracy to competing methods while requiring significantly less storage. In addition, we evaluate our ensembles in terms of predictive calibration and uncertainty, showing they compare favourably with the state-of-the-art. Finally, we draw a link with the continual learning literature, and we propose a modification of our framework to handle continuous streams of tasks with a sub-linear memory cost.
			We compare with a number of alternative strategies to mitigate catastrophic forgetting, highlighting advantages in terms of average accuracy and memory.
		\end{abstract}
		
    \end{frontmatter}

    
	\section{Introduction}
	An ensemble is the combination of multiple models, trained to solve the same task and \changemarker{joined} to improve the overall performance. Informally, in a good ensemble each member is accurate on its own, but makes independent errors during the prediction \cite{perrone1992networks}. By combining the outputs of its components, an ensemble can achieve better performance than any base member. For neural networks, a simple strategy for building an ensemble of models is to train repeatedly the same architecture starting from different initial conditions, and averaging the resulting predictions. In this setup, diversity is provided by the exponentially growing number of possible minima \cite{kawaguchi2016deep}, while averaging provides a degree of smoothing against minima with poor generalization capability.

	While this \textit{naive} strategy has been shown to provide significant advantages (e.g., \cite{lakshminarayanan2016simple,kroghneural,ovadia2019can,gustafsson2020evaluating}), it is not widely used in practice. The computational time required to train multiple networks and the memory cost needed to save them
	limit its use. 
	In fact, each model must be trained independently and stored in the system, and this can be infeasible when a single model is composed by millions of parameters, or when the hardware capacity is limited (e.g., mobile devices). 
	
	With the emergence of highly over-parameterized models, a number of alternative techniques, e.g., \cite{huang2017snapshot,wen2020batchensemble}, have been proposed to mitigate the space and time constraints. Key to many proposals is that, for a sufficiently large model, different components of the ensemble can share portions of the original network, opening the possibility of a sub-linear scaling of the memory with respect to the size of the ensemble \cite{wen2020batchensemble}. We briefly review these proposals in Section \ref{subsec:ensembles}. In general, one can observe a trade-off between the amount of sharing (i.e., the memory cost) and the performance of the ensemble. Whether this gap can be closed remains an open research question motivating this paper.
    	
	Ensembles of deep neural networks (called \textit{deep ensembles} in the following) are of particular interest in the field of continual learning (CL). In CL, a single network must solve multiple tasks sequentially, without forgetting already learned information \cite{thrun1998lifelong}. This property is hard to achieve and yet crucial for real-world scenarios, where the domain can change. When this is not possible and some, or all, the past learned information is lost, we have a phenomenon called catastrophic forgetting (CF, \cite{mccloskey1989catastrophic, french1999catastrophic}). Because of its importance, many algorithms were proposed recently to achieve effective CL with neural models. 
	Among these techniques (briefly reviewed in Section \ref{subsec:continual_learning}), several solutions find sub-structures within the original network, each one capable of solving a single task while leaving space for future tasks; usually, this is done by masking a portion of the network that is not used. 
	In this sense, these methods can be seen as an ensemble of sub-networks, each one specialized on a single task. Thanks to this neat separation, many of these methods are capable of mitigating CF \cite{parisi2019continual}.
	
	
	Deep ensembles and CL are problems that need to be addressed in order to achieve stable and robust agents that can operate autonomously in real-world applications and, as described before, methods developed in the two fields share a number of similarities. In this paper, we first develop in Section \ref{sec:proposed_method} a framework that is used for reducing the memory footprint of deep ensembles, without sacrificing the performances. Differently from other proposals in the literature, we optimize the members of the ensemble at \textit{initialization}, by solving an optimization problem defined over partial sub-masks of the original network. Then, we extend the method in Section \ref{sec:method_cl} for CL scenarios. We describe our contributions in more detail next.
	
	\subsection{Contributions of the paper}
	\subsubsection{Structured Ensembles}
	To build an ensemble of $N$ networks, we apply $N$ scaling vectors over the output of each inner and trainable layer in the original model. Each scaling vector is associated to a component of the ensemble. These vectors are trained by considering the network at initialization, while the networks' weights remain unchanged. In the end, a scaling value is used to estimate how important the associated neuron is for the given component of the ensemble. Only the most important ones are collected and the sub-network extracted, ignoring the others. This method is well motivated by the fact that, usually, a network is over-parameterized, and contains multiple sub-networks that can be extracted by its structure \cite{ramanujan2020s}, as demonstrated by the recent Lottery Tickets Hypothesis \cite{frankle2018lottery} and by several structured pruning strategies \cite{he2017channel}. 
	
	The overall procedure, which can be seen as a discovery of hidden paths in the original neural network, is the following: 1) the \changemarker{scaling vectors} are applied and trained for a few epochs, 2) for each component $i=1\dots N$ we collect the \changemarker{scaling vector} at position $i$ in each layer of the original network, 3) the neurons which are important for the current sub-network are kept, and the others ignored.
	The process produces $N$ smaller networks, and the dimension of each one can be regulated by choosing the percentage of how many neurons must be discarded in the extraction phase. An important property is that each network produced using this process is capable of achieving good performance despite the reduced number of parameters (even with a large percentage of discarded neurons), and when combining all the extracted networks into an ensemble the achieved score overcomes the one obtained using only the original network, due to the diversity of the networks that compose the ensemble.
	In order to create sub-networks that are all different from each other, we force a level of dissimilarity between \changemarker{scaling vectors} during training using a maximum mean discrepancy (MMD) regularization term \cite{pomponi2020bayesian}. To the best of our knowledge, the proposed method is the first which is capable of extracting a set of different sub-networks from a single untrained one. 
	
	We evaluate the efficiency of the proposed \textit{\nome{}} algorithm over different choices of architectures and datasets, and by investigating its robustness under different percentages of kept units. Moreover, we show that the ensembles created using \nome{} can be better calibrated than other approaches and the performances remain good when it comes to evaluating the network on out-of-distribution samples. \changemarker{We also perform a number of ablation studies on the algorithm, including over the choice of initialization strategy for the scaling vectors.}
	\subsubsection{Continual Learning}
	As the second main contribution of the work, we extend \nome{} to devise a novel approach for mitigating CF. In the CL scenario, the key idea is to associate each member of the ensemble (i.e., using a binary mask $m_i$) to a single task, by finding which portion of the network is maximally responsible for the performance on that task. In this way, we can remove CF by freezing these weights during the training of future tasks.
	
	The overall procedure is the following: 1) before solving a task, we identify a smaller sub-network that can be used to achieve good performance, but without extracting it, 2) during the training only the weights in this subset are modified, while the others are masked to zero, or used but not changed if they are part of a sub-network associated to a past task. 
	In this way, when we want to classify a sample from a given task, only the weights up to that task are used, and the others ignored. 
	
	In the experiments, we evaluate empirically how \nome{} performs on different CL scenarios,
	showing that the final accuracy achieved outperforms the other methods and that the additional memory required to store the binary masks is negligible; in the end, we show also how the hyper-parameters affect the final results, by performing different ablation studies. 
	
	
	\section{Related Work}
	In this section, we provide a short background on related works concerning ensembling (\ref{subsec:ensembles}), continual learning (\ref{subsec:continual_learning}), and structured pruning (\ref{subsec:structured_pruning}).

	\subsection{Ensembles}
	\label{subsec:ensembles}
	Deep ensembles \cite{lakshminarayanan2016simple}, originally inspired by bagging \cite{breiman1996bagging}, are models which combine multiple neural networks trained from randomly initialized weights.
	Ensembles have been theoretically studied to understand how and why the performances improve, as well as how to reduce the memory footprint and how to speed up the training time, often focusing on one of the two aspects. 
	Regarding the memory aspect, the networks can be distilled into smaller ones, without loss of performance \cite{hinton2015distilling}, or compressed, as exposed in \cite{bucilua2006model}. Regarding the training time aspect, instead, recently in \cite{huang2017snapshot} the authors proposed Snapshot Ensemble, in which a single model is forced to explore multiple local minima using cyclic learning rate schedules; the models obtained in these minima are saved and combined to build the final ensemble. Also, to reduce the training time, in \cite{garipov2018loss} the geometric properties of a loss function are used to connect different minima and extract multiple models, used to build the final ensemble. 
	
	In addition to these methods, some techniques to improve at the same time memory footprint and training time have been proposed. Recently, in \cite{wen2020batchensemble} the authors proposed Batch Ensemble, a method which uses the Hadamard product to change the weights and the output of each layer using multiple sets of trainable \changemarker{scaling vectors}, each one associated to a different ensemble's component. \changemarker{Similarly, in \cite{lee2015m}, the authors proposed a family of tree structured networks, called TreeNet, that simulates an ensemble by building a network consisting of zero or more shared initial layers, followed by a branching point and zero or more independent layers}. Another approach is called MC-Dropout, which uses dropout to simulate multiple networks, by dropping the weights also during the forward step in the inference phase \cite{gal2016dropout}. Both approaches are relaxations of the original ensemble formulation, and there is usually a deterioration of the performances (if compared to the original deep ensemble models). 
	
	Recently, some studies provide more insight into the deep ensemble approach, studying why and how this method performs better than others. In \cite{allen2020towards} the authors found that ensemble in deep learning works very differently from traditional learning theories, and, to understand and explain the hidden mechanisms of deep ensembles, they develop a new theory, while in \cite{fort2019deep} the ensemble approach is studied from a loss landscape prospective by measuring the similarity between the different models which compose the ensemble. 
	
	Other important aspects, that justify studying the ensemble approaches, are that these models are usually more calibrated than a single neural network \cite{guo2017calibration} and that it is possible to calculate a measure of uncertainty associated to a prediction. Both aspects are crucial in real-world applications, e.g., driver-less vehicles
	and medical applications \cite{kwon2020uncertainty}.

	\subsection{Continual Learning}
	
	\label{subsec:continual_learning}
	A continual learning (CL) scenario is one in which a model is trained to solve a number of sequential tasks
	, and should not forget how to solve past tasks while learning how to solve the current one \cite{thrun1998lifelong}. When the models lose the ability to solve past tasks
	we have a phenomenon called catastrophic forgetting (CF \cite{mccloskey1989catastrophic, french1999catastrophic}). A CL method should be capable of alleviating, or removing, CF while solving 
	efficiently the current task. 
	
	A CL scenario is a set of tasks and rules on how these tasks are acquired and which information can be retrieved from them. Many scenarios have been formulated, and we follow those proposed in \cite{van2019three}. More specifically, in this paper we focus on the scenario called Task-IL, where a task's identity is given during both training and testing, and the tasks contain disjoint classes. A generic model used to solve this kind of scenario is composed of a backbone network, that is shared between all tasks and therefore may suffer from CF, and a set of smaller networks, called \textit{solvers}, one per task, that are used to classify only the samples from the associated task, that are trained on top of the backbone. Despite its apparent simplicity, many approaches have been developed to solve this scenario. 
	
	
	Following the categorization proposed in \cite{parisi2019continual}, the methods to mitigate or remove CF in this scenario can be divided into three broad categories: \textbf{Rehearsal strategies}, \textbf{Regularization techniques}, and \textbf{Architectural strategies} (our proposal falls among the architectural category, on which we focus more).
	
	The \textbf{Architectural strategies} group contains all the methods that operate over the architecture of a neural network (e.g. custom layers, weights freezing/pruning) to mitigate or remove CF. Even if this group of methods has received less attention if compared to the others, many interesting approaches exist. In \cite{rusu2016progressive} the authors proposed Progressive Neural Networks, a model which is capable of expanding its size when a new task is encountered. It removes CF, but the required memory can be prohibitive. 
	Batch Ensembles \cite{wen2020batchensemble}, mentioned before, computes a set of \changemarker{scaling vectors} associated to the different tasks, training each of them independently. This approach requires small additional memory, but it is limited because the backbone is trained only during the first task, while for the other tasks the backbone is fixed and only the additional vectors are trained, limiting the final performance. 
	Another interesting approach, proposed in \cite{golkar2019continual}, is based on the classic pruning scheme, in which the most important weights for solving a task are saved and freezed, while the others are used to solve new tasks. The main problems of this method are that it is not possible to automatically decide the pruning percentage, and some accuracy can be lost when pruning the final model; also, past information is not used, limiting the achieved performance. 
	The closest method to the one we propose here is Supermasks In Superposition \cite{wortsman2020supermasks}, where for each task a binary mask over the backbone is learned. Our method has some similarities with these methods, but it is deeply different as a whole. In fact, it can be seen as a structured pruning approach, in which the masks are retrieved by searching for the best structure, instead of pruning the redundant weights after the training, and each structure is well separated from the others but reused to extract useful information from future samples, without compromising the results on past tasks.  

	\subsection{Structured Pruning}
	\label{subsec:structured_pruning}
	
	Our proposal has a strong connection with structured pruning approaches \cite{he2017channel, li2016pruning}. These methods do not prune each weight singularly, but entire groups of them together (e.g., neurons,  filters,  or  channels). Usually, these methods add a regularization term to the optimization process. Then, when the minimization is over, the groups are pruned. Our proposal, instead, is capable of pruning the neurons before the training process. To this end, we extend the SNIP approach from \cite{lee2018snip}, proposing a method that is capable of detecting multiple structured patterns within a neural network, used together to build up an ensemble of models, that is capable of solving efficiently a classification problem, without loss of performances. 
	
	\section{Proposed Method: \nome{}}
	\label{sec:proposed_method}
	
	As described before, deep ensembles of multiple neural networks suffer from an expensive memory footprint. In this section we introduce our method, which aims to reduce the required memory by creating smaller networks from an untrained one. 
	Firstly, we describe how the main idea works for reducing the dimension of a network. Secondly, we extend the method so that it can be used to extract multiple smaller networks within a single pre-processing procedure. 
	
	\subsection{Extracting a sub-network}
	\label{sec:extraction}
	\begin{figure}[t]
		\centering
		\includegraphics[width=0.8\linewidth]{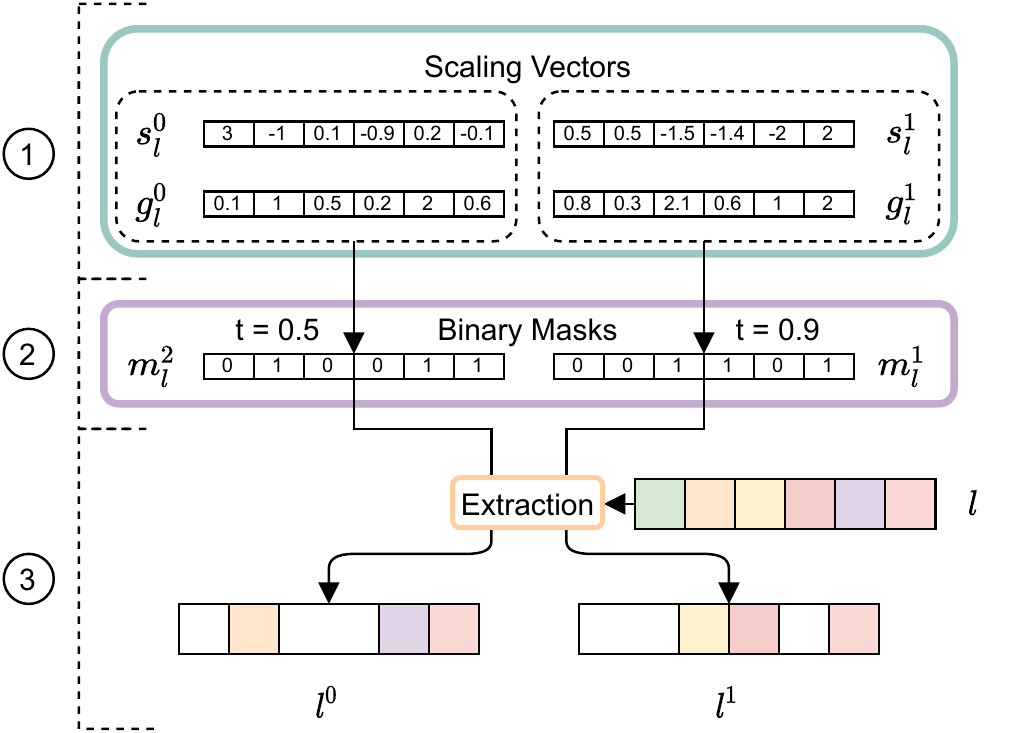}
		\caption{The image shows the extraction process of two linear layers from a single one $l$, using pruning percentage $p=0.5$. For simplicity, instead of showing the whole matrix of weights, only the output neurons of the layer are shown. The process starts in the step $1$: after the minimization step, the gradients associated to the \changemarker{scaling vectors} are collected. In the step $2$ the gradients are used to calculate the thresholds $th$, one for each \changemarker{scaling vector}, and thus the sets of indexes are created. These sets are used in the last step $3$, to create two new layers $l^0$ and $l^1$, by extracting the neurons in the original layer $l$ associated to indexes in the sets. 
		}
		\label{fig:method}
	\end{figure}
    
	Let us consider a neural network which is composed by $\text{L}$ layers $\mathbf{L}_l$ with $l=1\dots \text{L}$
	. To extract a substructure we want to consider only the layers that modify the dimension of its input, such as linear or convolutional layers. For example, let $\mathbf{L}_l$ be a generic linear layer with input size $i$ and output $o$ (the same can be done with a convolutional layer), that has weights matrix $W^l \in \mathbb{R}^{i \times o}$, and may have bias vector $b^l \in \mathbb{R}^{o}$, otherwise $b^l = 0$. Given an input $x \in \mathbb{R}^{i}$, we have the corresponding output $z^l = \mathbf{L}_l(x) = W^{l^T}x + b^l$. 
	
	Our intuition is that a matrix of weights contains a subset of the output neurons that can be extracted, and used to create a new layer with reduced dimensions. 
	To estimate the importance of an output neuron in a layer $\mathbf{L}_l$, we first multiply the output $z^l$ by a learnable real valued vector, called \changemarker{scaling vector},
	$s^l \in \mathbb{R}^{o}$, so that the new output is given by $\overline{z}^l = \mathbf{L}_l(x)\, \odot \, s^l$. 
	We initialize $s^l$ from the normal distribution $\mathcal{N}(0, 1)$. 
	
	
	
	Denote by $S = \bigcup s^l$ the union of all \changemarker{scaling vectors} through the network, and by $f(x; S)$ the output of the original network where every layer is replaced by its scaled version $\overline{z}^l$. We can find $S$ by solving an optimization problem over the \changemarker{scaling vectors}:
	\begin{equation}
		\min_S \mathbb{E}[ \mathcal{L}(y, f(x; S) ]
		\label{eq:opt_S}
	\end{equation}
	where $\mathcal{L}$ is the loss function for our problem, e.g., cross-entropy. We found that \eqref{eq:opt_S} requires only a small number of epochs to train the \changemarker{scaling vectors} to a reasonable value. Thus, it can be trained significantly faster than the original optimization problem over the weights of the network, which is necessary for keeping the approach bounded in terms of computational time.
	
	
	Once we have the set $S$, the \changemarker{scaling vectors} are used to measure how much each output neuron is important for solving the original problem. Directly using the final scaling values in a \changemarker{scaling vector} $s^l$ is not a good choice, since it does not take into consideration the correlation between the weights of a layer and its \changemarker{scaling values} (higher \changemarker{scaling values} do not imply more importance). To do this we need another estimator of this importance, and for \nome{} we use the one proposed in \cite{lee2018snip}. Given an input sample $x$ and its associated label $y$, the loss function $\mathcal{L}(\cdot, \cdot)$ and a \changemarker{scaling vector} $s^l$ associated to a linear layer $\mathbf{L}_l(x)$, we calculate the gradient associated to a scaling value $j$ as: 
	
	\begin{equation}
		\label{eq:global_gradient}
		g^l_j = g(s^l_j) = \frac{\Big\vert \frac{\partial \mathcal{L}(y, f(x; S))}{\partial s^l_j} \Big\vert}{ \sum_k^\text{L} \sum^N_i \Big\vert \frac{\partial \mathcal{L}(y, f(x; S))}{\partial s^k_i} \Big\vert}
	\end{equation}
	
	The gradient value $g(\cdot)$ can be computed using the loss over a single sample, a batch of samples, or the averaged gradients computed over the whole dataset. Differently from \cite{lee2018snip}, in \nome{} we use the value $g(\cdot)$ on the whole dataset, to avoid sub-optimal extraction of the most important neurons. 
	
	
	Once all the values $g(\cdot)$ are computed, we collect them into a set $G = \bigcup_l^\text{L} \{g(s^l_i) \,  \vert \, s^l_i \in s^l \}$. 
	To extract a sub-structure we need to define a threshold over the values in $G$: if a value $g(s^l_i)$ is higher than the threshold we want to keep the neuron $i$ of the layer $l$ also in the extracted network, otherwise it can be discarded. In \nome{}, the user sets the percentage of neurons $p$ to discard (as a hyper-parameter), 
	and the threshold is automatically calculated over the whole set of values as $th = Q_G(p)$. The function $Q_A(p)$ returns the percentile associated to the $p\%$ value of the set $A$. Once the threshold is calculated, we need to iterate over the layers of the network and extract the most important neurons. To do this, define the set of indexes associated to important neurons in the layer:
	
	
	\begin{equation}
		\label{eq:index_set}
		I_l = \{i \vert g^l_i > th \} \qquad i = 1 \dots o
	\end{equation}
	
	\noindent for each layer in the network $l$, where $o$ is its output size. 
	%
	Then, if the layer $l$ is the first one ($l = 1$), the new layer $\overline{\mathbf{L}}_l$ has a new weights matrix that is defined as $\overline{W}^l \in \mathbb{R}^{i \times \vert I_l \vert}$, where the following equivalence holds $\overline{W}^l_{j, k} = W^l_{j, k}$ for each $j=1 \dots i$ and for each $k \in I_l$. 
	The equivalence means not only that the weights have not been changed during the training of the \changemarker{scaling vectors}, but also that the input size of the layer is unchanged, while the output's size is composed only of the neurons that have been selected to be part of the new layer; the same process must be applied to the bias vector, if present. \changemarker{The extraction procedure for a linear layer is  visualized in Fig. \ref{fig:method}}.
	
	When the layer is not the first one, in addition to the exposed procedure, we extract also the input neurons associated to the selected output's neurons of the previous layer using the set $I_{l-1}$, further reducing the dimension of the layer $l$. The extraction process is the same used before, but applied to the input ones. 
	
	The process is iterated for all layers whose dimensions depend on the precedent, or subsequent, layer. In this way, changing a layer may have repercussions also on connected layers, further reducing the number of parameters in the extracted neural network. 
	
	To conclude, the set of new layers $\overline{\mathbf{L}}_l$ compose the new neural network $f_0(x)$, which has the same number of layers and connections between them, but fewer weights. 
	
    \begin{algorithm}[t]
    \changemarkertable{
    \caption{\changemarker{Structured Ensemble (pseudo-code).}}
    \label{algo:proposed_approach}
    \hspace*{\algorithmicindent} \textbf{Input} untrained network $y = f(x)$, a dataset $\mathcal{D} = \{(x, y)\}$, the number of subnetworks to extract $N$, and a pruning percentage $p$.\\
    \hspace*{\algorithmicindent} \textbf{Output} a set of networks $E$.
    \begin{algorithmic}[1]
    \State Initialize the masks $S_i$, for $i=1\dots N$ and create  $S = \bigcup S_i$.
    \State Solve the minimization problem: 	
    	\begin{equation*}
    			\min_{S}  \ \mathbb{E}[ \mathcal{L}(y, f(x; S) ] +
    			\frac{2\lambda}{(N(N-1))}
    			\sum_{i=1, j>i}^N R(S_i, S_j)^{-1}.
	    \end{equation*}
    
    \State $E$ = \{\}
    \For{$i =1 \dots N$}
    \State Retrieve $S_i$
    \State Given the dataset $D$, calculate the gradients $g^l$ for each mask in $S_i (Eq. \ref{eq:global_gradient})$.
    \State Calculate the threshold $th = Q_G(p)$ and $I_l$ for each layer $l$ (Eq. \ref{eq:index_set}).
    \State Extract the sub-network $\bar{f}(\cdot)$ of $f(\cdot)$, composed by the neurons that have $I_{i, l} = 1$.
    \State $E \leftarrow E \ \bigcup \ \{\bar{f}(\cdot)\}$.
    
    \EndFor
    \State Return the ensemble of networks $E$.
    \end{algorithmic}}
    \end{algorithm}

	\subsection{Extracting multiple sub-networks via diversity regularization}
	\label{sec:div_reg}
	
	The process described up to this point works for extracting one sub-network, but we want to extract $N$ sufficiently diverse sub-networks to build up an ensemble model. 
	To this end, we use a regularization term $R(\cdot, \cdot)$ to promote the distance between different \changemarker{scaling vectors} in a layer, and thus the extracted networks. 
	
	
	Instead of applying a single \changemarker{scaling vector} $s^l$ on a generic layer $l$, multiple vectors $s^{l, i}$ are applied, with $i=1 \dots N$, where $N$ is the number of networks to be extracted. To speed up the pre-processing, we train all the \changemarker{scaling vectors} in parallel, following the idea proposed in \cite{wen2020batchensemble}. Given a mini-batch of $B$ samples, we partition it into $B/N$ subsets, and apply a different set of scaling vectors to each subset. If $B$ contains fewer samples than $N$ or its dimension cannot be divided in $N$ equally sized sets, we replicate some input samples to make up the difference, in such a way that a \changemarker{scaling vector} does not see a sample twice in the same optimization step.
	
	
	Denoting by $S_i = \bigcup s^{l, i}$ the set of \changemarker{scaling vectors} associated to the $i$-th component, and by $S = \bigcup S_i$ the set of all the \changemarker{scaling vectors}, we can find $N$ sub-networks in parallel by solving:
	\begin{equation}
		\label{eq:reg_minimization}
		\begin{alignedat}{1}
			\min_{S} & \ \mathbb{E}[ \mathcal{L}(y, f(x; S) ] +\\
			& \frac{2\lambda}{(N(N-1))}
			\sum_{i=1, j>i}^N R(S_i, S_j)^{-1}
		\end{alignedat}
	\end{equation}
	where $R(\cdot, \cdot)$ is a regularization term forcing a desired level of diversity between a pair of masks, $\lambda$ is a hyper-parameter balancing the two terms, and $f(x; S_i)$ is the appropriately scaled version of $f$. 
	In particular, we regularize the training by applying the MMD distance \cite{Gretton:2012:KTT:2188385.2188410} between pairs of vectors in a layer. Since \changemarker{ the MMD distance} is symmetric, the comparisons are reduced to $\frac{N(N-1)}{2}$ for each layer. The resulting distances are summed, weighted, and added to the loss which is minimized during the training of the \changemarker{scaling vectors}. Formally, given a layer $l$ and two \changemarker{scaling vectors} $s^{l, i}$ and $s^{l, j}$, we calculate the distance as:
	
	\begin{equation*}
		\label{eq:mmd}
		\begin{alignedat}{1}
			\text{MMD}^2_\kappa(s^{l, i}, s^{l, j}) &= \frac{1}{n(n-1)}\sum_{z \ne k} 
			\kappa(s^{l, i}_z, s^{l, i}_k) \\ 
			&+ \frac{1}{n(n-1)}\sum_{z \ne k} \kappa(s^{l, j}_z, s^{l, j}_k) \\
			&- \frac{2}{n^2}\sum_{k, z} \kappa(s^{l, i}_z, ^{l, j}_k)
		\end{alignedat}
	\end{equation*}
	
	\noindent where $n$ is the dimension of the scale vector and $\kappa$ a kernel function. In the experiments we use the RBF kernel $\kappa(a, b) = \text{exp}\big\{\frac{\lVert a - b\rVert^2}{n}\big\}$. 
    
	After the training of the \changemarker{scaling vectors}, extracting the sub-networks proceeds as exposed before (Section \ref{sec:extraction}), but it is done $N$ times: for each $i= 1 \dots N$, we 
	use the set $S_i$ to calculate the neurons to keep, we extract the associated sub-network $f_i(x)$ and \changemarker{we save it} in the ensemble model. During this process the original network is not changed, because this would compromise future extractions. Figure \ref{fig:cl_method} shows the extraction process of two new layers from a generic linear layer $l$.
	
	The final ensemble is the set $E = \{f_i \vert i= 1 \dots N\}$, whose predictions can be combined in standard ways (e.g., the average for regression and a majority vote for classification). \changemarker{The complete procedure to extract multiple subnetworks from an untrained one, while regularizing the distance between the masks, is summarized in Algorithm \ref{algo:proposed_approach}.}
	
	

	\subsection{\nome{} as an approach to Continual Learning}
	\label{sec:method_cl}
	\begin{figure}[t]
		\centering
		\includegraphics[width=0.8\linewidth]{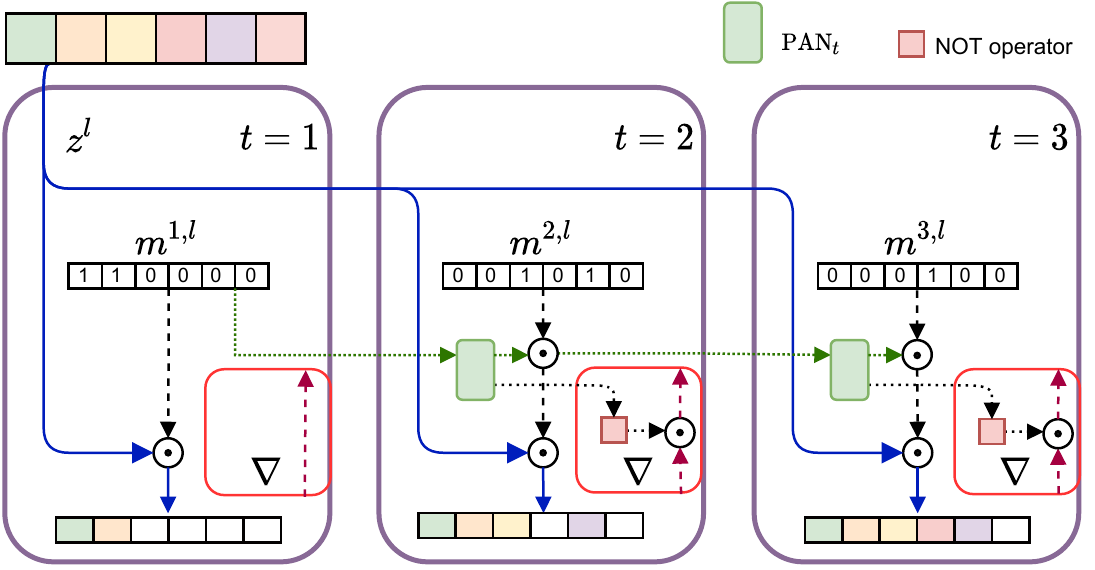}
		\caption{The image shows how the masks are combined to mask the output $z^l$ produced by a generic linear layer $l$. We follow the process for three tasks $t$, showing which output neurons and gradients are masked to avoid the changing of weights important to past tasks. In the first one the output vector is masked only using $m^{1, l}$, and the gradients $\nabla$ are unmodified. In the following tasks, past masks are used to build the $\text{PAN}_t$ set, that is combined with $m^{t, l}$ to mask the output of the layer, while the gradients are masked using the negation of the masks in $\text{PAN}_t$. }
		\label{fig:cl_method}
	\end{figure}
	%
	
	As said before, we use the Task-IL CL scenario, proposed in \cite{lopez2017gradient}. In this scenario we receive a sequence of disjoint tasks $t=1, \dots, M$, where a new task can be collected only when the current one is over. Each task is composed by a set of triples $\{(x_i, y_i, t)\}_{i=1}^S \in \mathcal{X} \times \mathbb{N^+} \times \mathcal{Y}$, where $x_i$ is a sample from the task (which is not present in other tasks), its label is $y_i$, which is defined with respect to the current task $t$, and $t$ is an index identifying the current task. We also define the neural network function $f(x; w, w_t)$, that given a sample $x$ produces the associated output $y$. The backbone is shared between all the tasks and its weights set is $w$, while the classifier head used to calculate the prediction depends on the task $t$ to which $x$ belongs, and its weights set is $w_t$. This separation is crucial, since in this scenario CL may arise only in the backbone network, and thus the weights $w_t$ in the solvers are not regularized by the methods.
	
	
	Our main intuition is the following: neurons trained on past tasks are capable of extracting information useful also for future tasks, and this can be exploited to use a smaller trainable structure, instead of the full network. In this way, a network is partitioned into logical smaller networks using binary masks, keeping weights used to solve past tasks unchanged (so that no CF occur), while solving new tasks. As in the ensemble approach, when we talk about neurons we refer to the output neurons of each inner layer of the backbone (linear or convolutional), in this way the binary masks are not applied on the weights, but only on the layers' outputs, further reducing the required memory. 
	
	When a new task $t$ is retrieved, we follow the same process for extracting a single network exposed before (Section \ref{sec:extraction}), but without extracting it. Instead, we define the set of Active Neurons of $t$ as $\text{AN}_t = \bigcup_{l=1}^{\text{L}} m^{t, l}$, where each binary mask $m^{t, l}$ is defined over the corresponding layer of the backbone as:  
	
	\begin{equation*}
		\label{eq:binary_mask_cl}
		m^{t, l}_i = \begin{cases}
			1 & \quad i \in I_l, \\
			0 & \quad i \notin I_l
		\end{cases}
	\end{equation*}
	
	\noindent where $I_{l}$ is calculated as in \eqref{eq:index_set}, using samples from task $t$, and the dimension of the mask is $m^{t} \in \mathbb{R}^{o}$. Also, this can be done globally, by calculating the masks using a globally calculated threshold, or locally to each layer, where a specific threshold is computed for each layer. When a set $\text{AN}_t$ is calculated, it must be saved in the memory, because it will be used also for all the tasks $t'>t$, as well as during the inference phase. These sets are the only additional memory required by the method.  
	
	If the task is the first one, $t=1$, the optimization problem to be solved is the following: 
	
	\begin{equation*}
		\min_{w, w_1} \mathbb{E}[ \mathcal{L}\left(y, f(x; w \odot \text{AN}_1, w_1)\right) ]
	\end{equation*}
	
	\noindent where $\odot$ is the element-wise product, and $w \odot \text{AN}_1$ means that, given the input sample $x$, and a layer $l$ with the associated output $z^l$, the latter is masked using the binary mask present in $\text{AN}_1$, accordingly to the layer's index, as: $\overline{z}^l = \mathbf{L}_l(x)\, \odot \, m^{1, l}$. The masks in $\text{AN}_1$ are used for zeroing the neurons which are not trained nor used during the training  of the first task. In other words, it forces the minimization process to train only the neurons that were extracted using the \changemarker{scaling vectors} associated to the current task, leaving the others unchanged.
	
	If the task is not the first, $t > 1$, we want also to use neurons associated to past tasks, without changing them. To do this, for a task $t$ and a layer $l$, we define the cumulative binary mask as:
	$$
	c(t, l) = \prod_{i=1}^{t-1} m^{i, l} 
	$$
	\noindent where each mask $m^{i, l} \in \text{AN}_i$. Then, we define a set called Past Active Neurons for the task $t$ as
	\begin{equation*}
		\text{PAN}_t = \bigcup_{l=1}^{\text{L}} c(t, l).
	\end{equation*}
	This set contains, for each layer, all the cumulative binary masks $c(t, l)$. In other words, it contains all the neurons which were active during past tasks, and \changemarker{that must be preserved} (not changed during the training), by masking the gradients associated to the weights which compose these neurons. At the same time, we want to use these neurons to improve the performance on the current task, by using the information they are capable of extracting. 
	
	Then
	, we need to extract the $\text{AN}_t$ set, and this must be done by taking into consideration also the masks in $\text{PAN}_t$. It can be done in two different ways: 
	\begin{itemize}
		\item Soft extraction: when calculating the threshold
		, also the neurons active in the $\text{PAN}_t$ set are taken into consideration. Practically, if a past active neuron is chosen, nothing changes in the \changemarker{final result}, since it belongs to $\text{PAN}_t$ anyhow and \changemarker{it must not be changed} in the minimization process. \changemarker{The dimension} of the $\text{AN}_t$ set is reduced, leaving more space for future tasks. This means that the procedure knows that some past neurons are more useful than others to solve the current task. 
		\item Hard extraction: in this case, the threshold and the extracted neurons are calculated only over the neurons which are set to $0$ in the $\text{PAN}_t$ set. 
	\end{itemize}
	
	\noindent Once we have these sets, we can minimize the optimization problem associated to the current task $t>1$. Using the same notation as before, the optimization process for the task $t$ is:
	\begin{equation*}
		\min_{w, w_t} \mathbb{E}[ \mathcal{L}\left(y, f(x; w \odot \text{AN}_t \odot \text{PAN}_t, w_t)\right) ]
	\end{equation*}
	\noindent To prevent the changing of the weights used by past tasks, we set to zero the gradients associated to these weights, and this is done at each step by multiplying these before the optimization step, as:
	\begin{equation*}
		w' = w - \eta \left[\frac{\partial \mathcal{L}(\cdot, \cdot)}{\partial w} \odot (\sim\text{PAN}_t)\right]
	\end{equation*}
	\noindent where $w'$ are the new weights, $\eta$ is the learning rate, and $\sim$ is the not operator, that negates all the  masks in $\text{PAN}_t$ ($\sim m^{t, l} = 1 - m^{t, l}$). The overall procedure is displayed in Figure \ref{fig:cl_method}.
	
	When we need to classify a sample $x$ from a task $t$, we follow the same steps exposed before: the $\text{AN}_t$ set is retrieved, if the task is not the first one $\text{PAN}_t$ is created, then, we use these sets to mask out the outputs of the layers while performing the forward step, using the function $f(x; w \odot \text{AN}_t \odot \text{PAN}_t, w_t)$. 

	\section{Experiments}
	In Section \ref{sec:classification} we compare \nome{} with different established ensemble methods, using multiple architectures and classification datasets. We evaluate different aspects of the ensemble models, such as accuracy and memory footprint (\ref{sec:exp_accuracy}), calibration (\ref{sec:exp_calibration}), and the capacity to detect out-of-distribution samples (\ref{sec:exp_diversity} and \ref{sec:exp_uncertainty}). In the end, different ablation studies are shown in Section \ref{sec:ablation}, in order to understand how the percentage of pruned neurons affects the final results.
	
	In Section \ref{sec:cl} we show the results on multiple CL benchmarks from the point of view of accuracy (\ref{sec:exp_cl_accuracy}) and memory (\ref{sec:exp_cl_memory}) required by each method. We conclude with a number of ablation experiments (\ref{sec:exp_cl_ablation}), in which we describe how the various hyper-parameters affect the final result on the CL benchmarks. 
	
	The code and all the files used to run the experiments are available online.\footnote{\url{https://github.com/jaryP/StructuredEnsemble}} 
    
	\subsection{Classification} 
	\label{sec:classification}
	\begin{table*}[t!]
		\centering
		\caption{Classification accuracy in percentage, on the classification benchmarks, using different architectures. The values are averaged over 3 experiments, and the standard deviation is also shown. Best results (among the ensembling approaches)
			for each dataset and network are highlighted in \textbf{bold}. Except for MC-Dropout \changemarker{(for which we do 5 sampling while testing)} and Single model, each method is an ensemble of $5$ neural networks. For TinyImagenet we used ResNet32 and VGG16; for the others dataset we used ResNet20 and VGG11. }
		\label{table:classification}
		\resizebox{0.8\textwidth}{!}{%
			\begin{tabular}{c|llcccc}
				\hline
				\multicolumn{1}{l}{Network} & \multicolumn{1}{l}{Method} & \multicolumn{1}{l}{SVHN} & \multicolumn{1}{l}{CIFAR10} & \multicolumn{1}{l}{CIFAR100} & \multicolumn{1}{l}{TinyImagenet}\\ \hline
				\multirow{6}{*}{ResNet} & Single model   &$94.08\scalebox{.8}{\rpm 0.72}$& $88.67\scalebox{.8}{\rpm 0.47}$ &$59.71\scalebox{.8}{\rpm 1.59}$ &$44.82\scalebox{.8}{\rpm 0.79}$\\
				& Deep Ensemble       & $96.24\scalebox{.8}{\rpm 0.11}$ & $92.35\scalebox{.8}{\rpm 0.05}$
				&$70.08\scalebox{.8}{\rpm 0.67}$ & $53.96\scalebox{.8}{\rpm 0.10}$ \\ \cline{2-6}
				& Snapshot       &$95.69\scalebox{.8}{\rpm 0.91}$& $90.38\scalebox{.8}{\rpm 0.08}$ & $61.84\scalebox{.8}{\rpm 0.11}$ &$48.75\scalebox{.8}{\rpm 0.53}$\\
				& Batch Ensemble &$94.57\scalebox{.8}{\rpm 0.72}$& $88.09\scalebox{.8}{\rpm 0.57}$ &$58.82\scalebox{.8}{\rpm 1.46}$ &$46.61\scalebox{.8}{\rpm 1.92}$\\
				& MC-Dropout     &$94.52\scalebox{.8}{\rpm 0.01}$&  $87.58\scalebox{.8}{\rpm 0.79}$ & $58.33\scalebox{.8}{\rpm 2.02}$ &$43.33\scalebox{.8}{\rpm 2.10}$\\
				& \nome{} (ours)           &$\mathbf{96.18\scalebox{.8}{\rpm 0.04}}$& $\mathbf{90.95\scalebox{.8}{\rpm 0.05}}$ &$\mathbf{67.55\scalebox{.8}{\rpm 0.66}}$&$\mathbf{49.72\scalebox{.8}{\rpm0.21}}$\\ \hline \hline
				\multirow{6}{*}{VGG}     & Single model   &$94.26\scalebox{.8}{\rpm 0.03}$&$88.67\scalebox{.8}{\rpm 0.47}$&$52.69\scalebox{.8}{\rpm 2.24}$ &$44.20\scalebox{.8}{\rpm 2.01}$\\
				& Deep Ensemble       &$95.80\scalebox{.8}{\rpm 0.06}$&$92.35\scalebox{.8}{\rpm 0.10}$&$60.18\scalebox{.8}{\rpm 0.29}$ &
				$52.45\scalebox{.8}{\rpm 0.60}$\\ \cline{2-6}
				& Snapshot       &$95.35\scalebox{.8}{\rpm 0.13}$&$89.38\scalebox{.8}{\rpm 0.08}$&$59.85\scalebox{.8}{\rpm 0.86}$ &$47.38\scalebox{.8}{\rpm 0.19}$\\
				& Batch Ensemble &$94.46\scalebox{.8}{\rpm 0.50}$&$88.09\scalebox{.8}{\rpm 0.57}$&$55.78\scalebox{.8}{\rpm 1.21}$ &$48.45\scalebox{.8}{\rpm 0.11}$\\
				& MC-Dropout     &$94.16\scalebox{.8}{\rpm 0.02}$&$87.58\scalebox{.8}{\rpm 0.79}$&$53.87\scalebox{.8}{\rpm 1.25}$ &$44.02\scalebox{.8}{\rpm 0.96}$\\
				& \nome{} (ours)           &$\mathbf{95.74\scalebox{.8}{\rpm 0.05}}$&$\mathbf{89.84\scalebox{.8}{\rpm 0.08}}$&$\mathbf{63.13\scalebox{.8}{\rpm 1.02}}$&$\mathbf{52.58\scalebox{.8}{\rpm0.42}}$\\ \hline
			\end{tabular}
		}
	\end{table*}
	%
	
	\-\hspace{0.25cm} \textbf{Datasets and architectures}: we evaluate \nome{} on different supervised classification datasets: CIFAR10 and CIFAR100 \cite{krizhevsky2009learning}, SVHN \cite{netzer2011reading}, and TinyImagenet (a subset of ImageNet \cite{Deng09imagenet} that contains $200$ classes). We test our method on different architectures, to study how the extraction of the sub-networks behaves. The architectures are ResNet-20 \cite{he2016deep} and VGG11 \cite{simonyan2014very} for, respectively, CIFAR10-100 and SVHN, while for TinyImagenet we used ResNet-30 \cite{he2016deep} and VGG16 \cite{simonyan2014very}. In this way we can test the extraction on sequential large networks such as VGG, and others with fewer parameters and with non-sequential architectures (due to the residual connections), like ResNet. Regarding the latter, applying the \changemarker{scaling vectors} on each layer is not possible, due to the residual connection. For this reason we apply the \changemarker{scaling vectors} on all the layers, except for the last one in each residual block. 
	In each experiment we use $5$ models to build each ensemble model. 
	
	\textbf{Baselines}: being our method an approach which aims to reduce the memory footprint of an ensemble model, we compare it with different established approaches that decrease the required memory or reduce the training time. For these reason we decided to compare it with \textbf{BatchEnsemble} \cite{wen2020batchensemble}, \textbf{Snapshot Ensemble} \cite{huang2017snapshot}, \textbf{MC-Dropout} \cite{gal2016dropout}, beside the \textbf{Single Model} approach (training a single neural network) and the standard ensemble of neural networks, which we call \textbf{Deep Ensemble}. These last two approaches are used only to provide a lower and upper baseline in terms of performance. \changemarker{We use the whole dataset for training each network, without performing any kind of bootstrapping.}
    
	\textbf{Hyper-parameters}: to set the hyper-parameters for each method we followed the respective papers. Regarding our method we test different percentages of pruning, and all obtained results are exposed in Section \ref{sec:ablation}. The best trade-off is obtained with hard extraction done layer-wise with percentage equals to $50\%$, and
	we train the \changemarker{scaling vectors} for $10$ epochs at the beginning of each task, setting $\lambda = 0.1$.   
	
	\textbf{Training procedure}: for each dataset and network combination we use the same training approach: we train for $200$ epochs using SGD optimizer with momentum equal to $0.9$, and starting learning rate equal to $0.01$ for VGG and $0.1$ for ResNet, that is annealed by multiplying it for $0.8$ every $50$ epochs.
	The only exception is Snapshot Ensemble, that we train for $50$ epochs per cycle. 
	Also, we extract $10\%$ of the training split to build up the evaluation dataset, and we use the accuracy score calculated on it to save the best model during the training, as well as the early stopping criterion (with tolerance $5$ for SVHN and $20$ for the other datasets). 
	
	We normalize each dataset so that the images pixels are in the range $[0, 1]$. Also, for CIFAR10-100 and SVHN we use the standard augmentation scheme, in which the images are zero-padded with $4$ pixels on each side and randomly cropped to produce $32\times32$ images, then horizontally mirrored with probability $0.5$; we use the same scheme also for SVHN and TinyImagenet, by scaling the cropped dimensions of the latter according to the sizes of its images.
	
	We repeat each experiment $3$ times, using incremental seeds from $0$ to $2$. 
	
	\subsubsection{Accuracy and memory}
	\label{sec:exp_accuracy}
	\begin{figure}[t]
		\centering
		\includegraphics[width=0.6\linewidth]{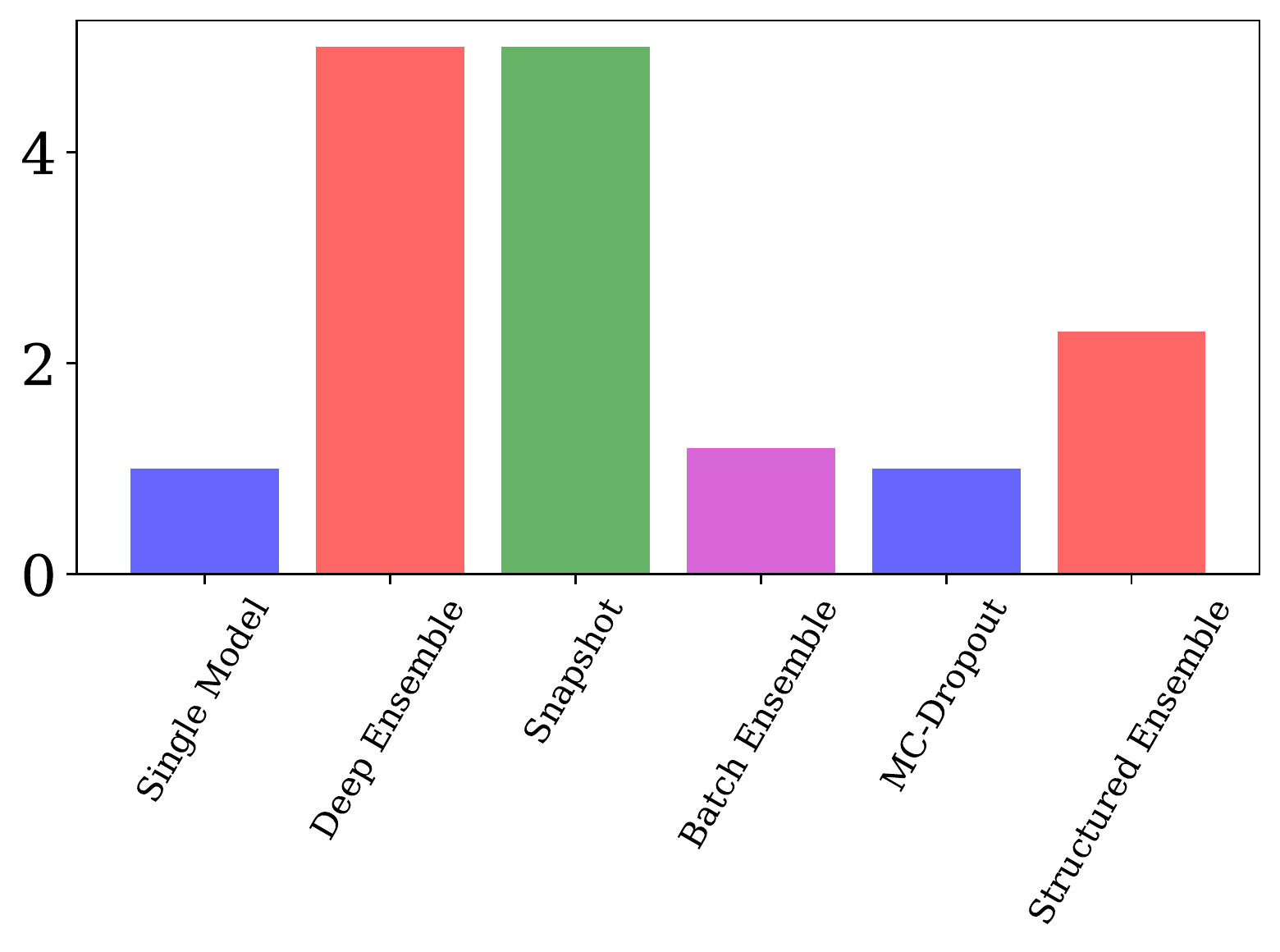}
		\caption{The image displays the overhead required by each method with respect to the base model, which requires 1. The results are associated to an ensemble of $5$ ResNet20.}
		\label{fig:memory}
	\end{figure}
	
	\label{sec:accuracy}
	%
	The accuracy results are summarized in Table \ref{table:classification}, while Fig. \ref{fig:memory} displays memory usage for each method \changemarker{in terms of the number of trainable parameters of the method over the number of trainable parameters required by a single neural network}, called Overhead (e.g. a deep ensemble model composed by $5$ networks has overheard equal to $5$).
	
	We can see, as expected, that all the models obtain a better, or similar, accuracy score than the single network, but worse than the naive ensemble approach. The methods that emulate an ensemble, Batch Ensemble and MC-Dropout, give the worst accuracy results, but require a small amount of memory (respectively close or equal to a single model). On the other hand, Snapshot and \nome{} require more memory but the achieved scores are close, even higher when using VGG on CIFAR100 or TinyImagenet, to the naive ensemble approach. In terms of memory, there is no difference between Snapshot Ensemble and the Deep Ensemble approach, while the memory used by \nome{} depends on the pruning percentage used (as we will see). 
	
	
	\subsubsection{Calibration}
	\label{sec:exp_calibration}
	\begin{table}[ht]
	\centering
	\caption{The resulting ECE (lower is better) scores obtained for all the methods using ResNet architectures (results associated to the ones displayed in Table \ref{table:classification}). The best results are highlighted in bold. }
	\label{table:ece}
	\resizebox{0.7\linewidth}{!}{%
		\begin{tabular}{l|c|c|c|}
			\cline{2-4} \cline{2-4}
			& CIFAR10 & CIFAR100 & TinyImagenet \\ \hline
			\multicolumn{1}{|l|}{Single model}  &$8.21\scalebox{.8}{\rpm 0.17}$&$19.60\scalebox{.8}{\rpm 3.97}$&$3.19\scalebox{.8}{\rpm 0.79}$\\ \hline
			\multicolumn{1}{|l|}{Deep Ensemble} &$1.61\scalebox{.8}{\rpm 0.04}$&$7.69\scalebox{.8}{\rpm 0.32}$&$1.56\scalebox{.8}{\rpm 0.72}$\\ \hline \hline
			\multicolumn{1}{|l|}{Snapshot}      &$\mathbf{3.01\scalebox{.8}{\rpm 0.21}}$&$12.93\scalebox{.8}{\rpm 0.43}$&$5.60\scalebox{.8}{\rpm 0.12}$\\ \hline
			\multicolumn{1}{|l|}{Batch Ensemble} &$8.14\scalebox{.8}{\rpm 1.24}$&$20.52\scalebox{.8}{\rpm 1.54}$&$2.73\scalebox{.8}{\rpm 0.14}$\\ \hline
			\multicolumn{1}{|l|}{MC-Dropout} &$7.11\scalebox{.8}{\rpm 0.12}$&$13.63\scalebox{.8}{\rpm 1.83}$&$22.54\scalebox{.8}{\rpm 0.87}$\\ \hline
			\multicolumn{1}{|l|}{\nome{} (ours)}&$3.91\scalebox{.8}{\rpm 0.30}$&$\mathbf{4.02\scalebox{.8}{\rpm 0.23}}$&$\mathbf{2.24\scalebox{.8}{\rpm 0.33}}$\\ \hline
		\end{tabular}%
	}
	\end{table}
	%
	The calibration of a network is well defined and can be easily calculated. Given a sample $x$, the associated ground truth label $y$, the predicted class $\bar{y}$ and its associated probability $\bar{p}$, we define the calibration of a method as: 
	\begin{equation*}
		\label{eq:calibration}
		p(\bar{y} = y \ \vert \ \bar{p} = p) = p \ \ \forall \ p \in [0, 1] \,.
	\end{equation*}
	\noindent This quantity cannot be computed with a finite set of samples, but, as proposed in \cite{niculescu2005predicting}, it can be approximated using a metric called Expected Calibration Score (ECE, lower is better), that measures how much the classification's  probabilities over a dataset are miscalibrated, by combining two different quantities.
	Before defining how to calculate the ECE, we need to choose the number $Z$ of equally sized bins in the range $[0, 1]$, and grouping a set of predictions into $Z$ interval bins $B_z$. Each $B_z$ groups together the samples with a prediction confidence $\bar{p}$ that falls into the range $( \frac{z-1}{Z}, \frac{z}{Z}]$. The ECE metric is calculated as:

	\begin{equation*}
		\label{eq:ECE}
		\text{ECE} = \sum_{z=1}^Z \frac{\vert B_z \vert}{n}  \Big| \text{acc}(B_z) - \text{conf}(B_z)\Big|
	\end{equation*}
	\noindent where $n$ is the number of samples, $\text{acc}(B_z)$ and $ \text{conf}(B_z)$ are, respectively, the averaged accuracy and the averaged confidence of the samples in $B_z$.

	In Table \ref{table:ece} the resulting ECEs associated to ResNet20 trained on CIFAR10 are displayed.
	The scores obtained depends heavily on the dataset type. For CIFAR10 we have that the best method is Snapshot Ensemble, and \nome{} is the second best; the other methods achieve a result that is similar to the Single Model, which is poorly calibrated. When the dataset becomes more difficult, Snapshot ensemble is one of the worst methods; we believe that this happens because the length of the learning cycles must be set accordingly to the dataset. On all the other dataset our proposal is the method that achieves the better ECE score. In the end, \nome{} is the approach that gives the most reliable results. 
	
	
	
	\subsubsection{Diversity analysis}
	\label{sec:exp_diversity}
		\begin{table}[t]
		\centering
		\caption{The diversity metrics obtained using all the methods with ResNet20 trained on CIFAR10 (results associated to the ones displayed in Table \ref{table:classification}).}
		\label{table:diversity}
		\resizebox{0.7\linewidth}{!}{%
			\begin{tabular}{l|c|c|}
				\cline{2-3} \cline{2-3}
				& CC Diversity & WC Diversity \\ \hline
				\multicolumn{1}{|l|}{Deep Ensemble} &$5.73\scalebox{.8}{\rpm 0.42}$&$32.89\scalebox{.8}{\rpm 1.56}$\\ \hline \hline
				\multicolumn{1}{|l|}{Snapshot}      &$5.60\scalebox{.8}{\rpm 0.12}$&$31.67\scalebox{.8}{\rpm 0.77}$\\ \hline
				\multicolumn{1}{|l|}{Batch Ensemble} &$2.73\scalebox{.8}{\rpm 0.14}$&$19.21\scalebox{.8}{\rpm 1.92}$\\ \hline
				\multicolumn{1}{|l|}{MC-Dropout} &$3.62\scalebox{.8}{\rpm 0.45}$&$22.54\scalebox{.8}{\rpm 0.87}$\\ \hline
				\multicolumn{1}{|l|}{\nome{} (ours)}&$3.99\scalebox{.8}{\rpm 0.70}$&$26.61\scalebox{.8}{\rpm 1.45}$\\ \hline
			\end{tabular}%
		}
	\end{table}
	The evaluation metrics used, such as accuracy and ECE score, do not express the key property of an ensemble model: the diversity of the models that compose it. In \cite{fort2019deep} the authors proposed to measure the diversity of two networks based on how many samples they classify in the same way. We use a slightly modified version, that measures the similarity using the probabilities predicted by each model in the ensemble.
	
	
	We define the function $g_i(x)$, which is associated to a network $i$, and, for a given sample $x$, it returns the unbounded prediction vector $v \in \mathbb{R}^c$, where $c$ are the classes of the dataset taken into consideration.
	Give this formulation, we define the diversity calculated over a set of samples $S$, as $\frac{1}{S}\sum_i^S \text{H}\left(\frac{1}{N}\sum_j^N g_j(x_i)\right)$, where $\text{H}(\cdot)$ is the function that transforms the input vector into a distribution using the soft-max function, and calculates the entropy of it, normalized in $[0, 100]$. Using this formulation we have that, for a single sample, the higher the entropy is, the more dissimilar networks' prediction are. 
	
	To evaluate the diversity of networks that compose an ensemble model, we calculate it on Correctly Classified (CC Diversity) samples and Wrongly Classified (WC Diversity), both calculated on the test set. In Table \ref{table:diversity} the results obtained using an ensemble of $5$ ResNet20 trained on CIFAR10 are displayed.
	
	We can see that, overall, each method gives low diversity for CC samples, and higher for WC ones. Also, Batch Ensemble and MC-Dropout have smaller difference between CC and WC samples, while Deep Ensemble and Snapshot ensemble display wider gap between CC and WC samples. Regarding \nome{}, we see that the predictions are sufficiently diverse, and the scores are similar to the one obtained by Deep Ensemble. 
	
	
	
	\subsubsection{Uncertainty}
	\label{sec:exp_uncertainty}
	\begin{table*}[t]
		\centering
		\caption{The results of the uncertainty used to discard the corrupted images. The results are associated to an ensemble model composed of $5$ ResNet20 trained on CIFAR10, using the trained networks obtained in Table \ref{table:classification}. The results are computed over the test split, which is composed by $10000$ images. We only show the results associated to a medium and high corruption of the images (the results associated to Corrupted CIFAR10 are averaged over all the types of corruption). The format of the results is the following $\text{A}\xrightarrow[]{\scalebox{.5}{\text{D}}}\text{FA}$, where the definition of A, D, and FA can be found in Section \ref{sec:exp_uncertainty}. This formulation means that the accuracy changes from A to FA by discarding D percentages of images with high associated uncertainty. For each corruption level we highlighted in bold the best Filtered Accuracy FA.}
		\label{table:uncertainty}
		\resizebox{0.9\linewidth}{!}{%
			\begin{tabular}{l|c|c|c|c|}
				\cline{2-5}
				\multirow{3}{*}{Method}                               & \multicolumn{2}{c|}{FGSM}  & \multicolumn{2}{c|}{C-CIFAR10} \\ \cline{2-5} 
				& \multicolumn{2}{c|}{$\epsilon$} & \multicolumn{2}{c|}{Severity}          \\ \cline{2-5} 
				&$0.02$&$0.5$&$3$&$5$\\ \hline
				\multicolumn{1}{|l|}{Snapshot}                          &$49.95\xrightarrow[]{\scalebox{.8}{45.62}}65.77$&$21.18\xrightarrow[]{\scalebox{.8}{48.07}}25.91$ &$68.25\xrightarrow[]{\scalebox{.8}{52.53}}86.12$&$50.12\xrightarrow[]{\scalebox{.8}{57.34}}71.02$\\ \hline
				\multicolumn{1}{|l|}{Batch Ensemble}                         &$42.00\xrightarrow[]{\scalebox{.8}{69.08}}65.39$&$21.17\xrightarrow[]{\scalebox{.8}{80.13}}32.10$ &$67.49\xrightarrow[]{\scalebox{.8}{54.45}}87.37$&$49.49\xrightarrow[]{\scalebox{.8}{69.97}}72.12$\\ \hline
				\multicolumn{1}{|l|}{MC-Dropout}                  &$48.39\xrightarrow[]{\scalebox{.8}{71.70}}76.89$&$22.06\xrightarrow[]{\scalebox{.8}{82.51}}38.42$ &$70.32\xrightarrow[]{\scalebox{.8}{52.83}}91.99$&$52.76\xrightarrow[]{\scalebox{.8}{67.36}}78.17$\\ \hline
				\multicolumn{1}{|l|}{\nome{} (ours)} &$48.06\xrightarrow[]{\scalebox{.8}{72.27}}\mathbf{78.57}$&$21.08\xrightarrow[]{\scalebox{.8}{86.56}}\mathbf{41.88}$ &$70.01\xrightarrow[]{\scalebox{.8}{57.21}}\mathbf{92.10}$&$51.27\xrightarrow[]{\scalebox{.8}{70.78}}\mathbf{82.12}$\\ \hline
			\end{tabular}%
		}
	\end{table*}
	Here we study how the methods behave when it comes to detect Out-of-Distribution (OOD) samples. Our main idea is to study if a method can detect and discard less confident predictions, improving the accuracy (if compared to the same prediction without discarding samples). 
	
	To do this we rely on two different experiments. In the first one we follow the same scenario proposed in \cite{pomponi2020bayesian}: we attack the ensemble models using Fast Gradient Sign Method \cite{goodfellow2014explaining} that, given an image $x$ and its label $y$, corrupt $x$ as $x + \epsilon \ \text{sign}(\nabla_x J(\theta, x, y))$, where $J$ is the input-output Jacobian and $\theta$ is the set that contains the parameters of the network. The idea is to understand if the model is capable of detecting whether an image has been attacked too much, making correct classification impossible. In the second experiment we evaluate the uncertainty on the recently proposed Corrupted CIFAR10 dataset \cite{hendrycks2019benchmarking}. The dataset consists of the same $10000$ images present in the test spit of CIFAR10, but corrupted with $20$ types of corruptions, each one with $5$ different severity levels.
	
	To understand if an image is corrupt and must be discarded we use the discarding process proposed in \cite{pomponi2020bayesian}. After a preliminary evaluation of the method, we decided to use the following formulation:
	
	\begin{equation*}
		\label{eq:thres}
		\text{T}_\gamma(\mathcal{H}) = \text{Q}_\mathcal{H}(75)
	\end{equation*}

	\noindent where function $\text{Q}_\mathcal{H}(75)$ returns the percentile associated to the $75\%$-th value of the set $\mathcal{H}$. In our experiments $\mathcal{H}$ is the entropy of each correctly classified image in the development set, calculated after the training process.
	
	Given a threshold and a corrupted set of samples that we want to filter by discarding highly corrupted images, we evaluate the performances by measuring different quantities:
	
	\begin{itemize}
		\item Accuracy (A): the accuracy calculated over the corrupted set. 
		
		\item Filtered Accuracy (FA): the accuracy calculated only over the images that have not been discarded. 
		
		\item Discarded samples (D): the percentage of images that have been discarded. 
		
	\end{itemize}
	
	\noindent A good method should be able to increment the accuracy by discarding images on which the confidence is low (maximize the difference between FA and A), while keeping the highest number of images on which the accuracy is calculated (minimize D). 
	
	
	In Table \ref{table:uncertainty} the results obtained using ResNet20 trained on CIFAR10 are exposed. To study how the performances are affected by the corruptions, we average the scores obtained on moderate and high corrupted images, by setting $\epsilon$ equals to $0.02$ and $0.5$ for FGSM, and using severity equals to $3$ and $5$ for C-CIFAR10. 
	
	
	We see that Batch Ensemble and Snapshot are the worst methods. In fact, the latter is not capable of discarding enough corrupted images (due to the similarity between the networks), while the first discards more samples than the others methods, but is not capable of reaching good scores if compared to the other methods; this means that the discarded samples also contain images that would have been classified correctly. These results are expected, because these methods are not capable of building an ensemble in which the models are different enough. Regarding MC-Dropout, despite being  proposed mainly for uncertainty estimation, it has lower performance than \nome{}, especially when the severity value is high, where it reaches a score that is $4\%$ higher than the second best one, while discarding a similar number of images.
	
	\subsubsection{Ablation Studies}
	\label{sec:ablation}
	\begin{table}[ht]
	\centering
	\caption{The resulting metrics obtained when changing the pruning percentage, averaged over 3 experiments (the standard deviation is also shown). The definitions of these metrics are in Section \ref{sec:classification}. The displayed results are associated to an ensemble model composed of $5$ ResNet20 trained on CIFAR10.}
	
	\label{table:pruning}
		\resizebox{0.8\linewidth}{!}{%
			\begin{tabular}{l|c|c|c|c|c|}
				\cline{2-6}
				& Accuracy & ECE & Overhead 	 & CC Diversity & WC Diversity \\ \hline
				\multicolumn{1}{|l|}{$30\%$} &$91.51\scalebox{.8}{\rpm 0.10}$&$4.54\scalebox{.8}{\rpm 0.26}$&$3.39\scalebox{.8}{\rpm 0.01}$&$2.85\scalebox{.8}{\rpm 0.41}$&$22.62\scalebox{.8}{\rpm 1.07}$\\ \hline
				\multicolumn{1}{|l|}{$50\%$} &$90.66\scalebox{.8}{\rpm 0.12}$&$3.91\scalebox{.8}{\rpm 0.30}$&$2.39\scalebox{.8}{\rpm 0.01}$&$3.99\scalebox{.8}{\rpm 0.70}$&$26.61\scalebox{.8}{\rpm 1.45}$\\ \hline
				\multicolumn{1}{|l|}{$60\%$} &$90.59\scalebox{.8}{\rpm 0.46}$&$3.61\scalebox{.8}{\rpm 0.23}$&$2.38\scalebox{.8}{\rpm 0.02}$&$3.88\scalebox{.8}{\rpm 0.36}$&$25.92\scalebox{.8}{\rpm 1.76}$\\ \hline
				\multicolumn{1}{|l|}{$70\%$} &$89.43\scalebox{.8}{\rpm 0.32}$&$2.78\scalebox{.8}{\rpm 0.25}$&$1.42\scalebox{.8}{\rpm 0.04}$&$5.87\scalebox{.8}{\rpm 0.26}$&$31.12\scalebox{.8}{\rpm 0.26}$\\ \hline
				\multicolumn{1}{|l|}{$80\%$} &$88.40\scalebox{.8}{\rpm  0.51}$&$1.55\scalebox{.8}{\rpm  0.24}$&$0.92\scalebox{.8}{\rpm 0.02}$&$8.21\scalebox{.8}{\rpm 0.27}$&$35.42\scalebox{.8}{\rpm 0.50}$\\ \hline
			\end{tabular}%
		}
	\end{table}
	In this section we study how the pruning percentage affects the results. Table \ref{table:pruning} contains all the results associated to the different pruning percentages studied. 
	
	\textbf{Accuracy and overhead}: from the table we can see that the accuracy decreases with the increase of pruned neurons, as expected, but it is always higher than the score obtained using a single network. Regarding the overhead, we can see that the additional memory required by the method is always lower than deep ensemble (which requires $5$ in this case), and when the pruning percentage grows, \nome{} produces an ensemble model that requires less memory than the base model, but achieves higher scores. 
	
	\textbf{Calibration}: the calibration of our proposal depends on the pruning percentage: we see that the calibration is poor when the pruning percentage is low, but, if we can sacrifice some accuracy points (around $2\%$), we can build up an ensemble which is better calibrated than all the other methods. 
	These results are expected, because the accuracy score and the ECE are deeply connected, and usually, a network loses the calibration of its predictions with the improving of the accuracy. In fact, ensembles composed by smaller networks gives slightly worst results (around $1-3\%$ of accuracy is lost) but better ECE. 
	
	\textbf{Diversity}: from the table we see that the CC diversity is always low regardless of the pruning percentage, but WC diversity depends on the pruned neurons, and it grows with the growing of the pruning percentage. This phenomenon can be explained thinking that smaller networks tend to be very different from each other, while bigger ones can converge to similar minima, and may give similar predictions. In fact, with the increase of the pruning percentage also the CC diversity increases, and with it the difference between CC and WC diversities.  
	
	\begin{table}[t]
	\centering
	\caption{\changemarker{The resulting metrics obtained when changing the initialization of the masks, using different distributions from which the values are sampled. The definitions of these metrics are in Section \ref{sec:classification}. The displayed results are associated to an ensemble model composed of $5$ ResNet20 trained on CIFAR10, build using \nome{} with pruning percentage set to $50\%$}}
	\label{table:initialization}
		\resizebox{0.8\linewidth}{!}{%
		\changemarkertable{
			\begin{tabular}{c|c|c|c|c|c|}
				\cline{2-6}
				& Accuracy & ECE & Overhead 	 & CC Diversity & WC Diversity \\ \hline
				\multicolumn{1}{|l|}{$\mathcal{N}(0, 1)$} &$91.05$&$3.80$&$2.39$&$3.34$&$24.22$\\ \hline
				\multicolumn{1}{|l|}{$\mathcal{N}(0, 2)$} &$90.93$&$4.11$&$2.39$&$3.34$&$24.22$\\ \hline
				\multicolumn{1}{|l|}{$\mathcal{N}(0, 5)$} &$90.87$&$3.92$&$2.48$&$4.16$&$26.02$\\ \hline \hline
				\multicolumn{1}{|l|}{$\mathcal{U}(0, 1)$} &$90.86$&$4.01$&$2.41$&$5.47$&$25.03$\\ \hline
				\multicolumn{1}{|l|}{$\mathcal{U}(-1, 1)$} &$90.74$&$3.94$&$2.30$&$4.74$&$29.11$\\ \hline
			\end{tabular}}%
		}
	\end{table}
	
	\changemarker{
	\textbf{Initialization of the masks}: in Table \ref{table:initialization} the results obtained when varying the initialization of the masks are shown. We can see that the obtained results do not depend heavily on the initialization of the masks, because these are used mainly to estimate the gradient value associated to each output neuron. However, when the variance becomes too large, we observe a slight decrease of the performances. In the end, we decided to initialize each mask using the standard distribution $\mathcal{N}(0, 1)$.
	}

	\subsection{Continual learning}
	\label{sec:cl}
	
	\-\hspace{0.25cm} \textbf{Continual learning setup}: any dataset can be used to build a CL dataset. To do this, we group different labels from the dataset into disjoint subsets, and, for each set of labels, we extract the samples associated to its components from the original dataset. In this way, we can split a dataset having $c$ classes into $M$ tasks, each one composed by $\frac{c}{M}$ classes. We use this established approach to build 3 CL benchmarks: 1) Split-MNIST, which contains $5$ tasks, each one with two classes, 2) Split-CIFAR10 \cite{rebuffi2017icarl}, with the same configuration of Split-MNIST, and 3) Split-CIFAR100, that contains $10$ tasks, each one with $10$ classes. For MNIST we used LeNet-5 \cite{lecun1989backpropagation}, while for the others datasets we used VGG11 \cite{simonyan2014very}, with the exception that we halved the size of each kernel. We trained all networks using Adam optimizer \cite{kingma2014adam}, with learning rate equal to $0.001$. To have a better statistic of the results we repeat each experiment $3$ times, as in the ensemble experiments. 
	
	\textbf{Baselines}: being our method a structural one, we compare it with similar approaches proposed in the recent years:  \textbf{BatchEnsemble} \cite{wen2020batchensemble} (also used in the ensemble experiments), \textbf{Supermask in Superposition} \cite{wortsman2020supermasks}, and \textbf{Pruning} \cite{golkar2019continual}. Also, we selected two other baselines, which are not CL methods: the first one is \textbf{Single Model} approach, in which we train a model for each task, and the second is \textbf{Naive}, where a single network is trained sequentially on the tasks without any method to alleviate CF. These last two can be seen as, respectively, upper and lower bounds for the scores achievable by the other methods; for this reason we do not compare the CL methods directly to them. 
	
	\textbf{Hyper-parameters}: to find the best hyper-parameters for each method we followed the respective paper. Regarding our method we test different combination of parameters, which are the same exposed before. We ended up with hard extraction done layer wise with percentage equals to $50\%$, and
	we train the \changemarker{scaling vectors} for $10$ epochs at the beginning of each task.   
	
	\textbf{Metrics}: 
	to evaluate the efficiency and to compare the CL methods usually two metrics are used. One shows the effective accuracy obtained on all the tasks, while the second indicates the amount of accuracy over past tasks that is lost while training new ones. In this case, the selected CL methods completely remove the CF phenomenon and for this reason we use only one metric from \cite{diaz2018don}, simply called Accuracy. It is calculated on a matrix $R \in \mathbb{R}^{M \times M}$, where $M$ is the number of tasks, and each entry $R_{ij}$ is the test accuracy on task $j$ when the training on task $i$ is completed. Given this matrix of scores, the accuracy is calculated as:
	\begin{align*}
		\text{Accuracy} = \frac{\sum_{i>j}^M R_{ij}}{\frac{1}{2} M(M+1)} \,.
	\end{align*}
	
	In addition to this metric, we compared the methods under the point of view of the memory required, by approximating the number of additional float or binary values used.
	

	\subsubsection{Classification results}
	\label{sec:exp_cl_accuracy}
	\begin{table}[t]
		\centering
		\caption{Final Accuracy averaged over all the tasks, evaluated when the last one is over, is shown. The values are averaged over 3 experiments, and the standard deviation is also shown. Best results for each dataset are highlighted in \textbf{bold}. Being the all methods structural, the CF phenomenon is not present (it is present only using the naive approach, but it is not shown).}
		\label{table:cl}
		\resizebox{0.8\linewidth}{!}{%
			\begin{tabular}{|l| c|c|c|}
				\hline
				Method         & S-MNIST & S-CIFAR10 & S-CIFAR100 \\ \hline \hline
				Naive          &$94.32\scalebox{.8}{\rpm 0.04}$&$69.99 \scalebox{.8}{\rpm 2.45}$&$29.39 \scalebox{.8}{\rpm 0.40}$\\ \hline
				Single Model  &$99.85 \scalebox{.8}{\rpm 0.04}$&$92.62 \scalebox{.8}{\rpm 0.82}$&$70.44 \scalebox{.8}{\rpm 0.15}$\\ \hline \hline
				SupSup        &$99.45 \scalebox{.8}{\rpm 0.02}$&$91.06 \scalebox{.8}{\rpm 0.11}$&$65.59 \scalebox{.8}{\rpm 0.34}$\\ \hline
				Pruning       &$97.91 \scalebox{.8}{\rpm 0.76}$&$78.23 \scalebox{.8}{\rpm 1.56}$&$49.27 \scalebox{.8}{\rpm 0.49}$\\ \hline
				Batch Ensemble &$99.34 \scalebox{.8}{\rpm 0.07}$&$85.40 \scalebox{.8}{\rpm 0.07}$&$56.12 \scalebox{.8}{\rpm 0.30}$\\ \hline
				\nome{} (ours) &$\mathbf{99.66 \scalebox{.8}{\rpm 0.26}}$&$\mathbf{92.27\scalebox{.8}{\rpm 0.32}}$&$\mathbf{66.24\scalebox{.8}{\rpm 0.39}}$\\ \hline
			\end{tabular}%
		}
	\end{table}
	In Table \ref{table:cl} the results are shown. 
	We can see that our method outperforms the others, since it is capable of achieving higher performances (while also requiring small additional memory, as we will see). This is accentuated when the complexity of the problem increases, starting from CIFAR10, in which our method  gives performances which are similar to the upper bound (Single Model baseline). We can conclude that our intuition about the usability of past information improves the performances, compared to the other methods that use a more neat separation of the sub-networks for solving the tasks. 
	
	\subsubsection{Memory usage}
	\label{sec:exp_cl_memory}
	\begin{table}[t]
		\centering
		\caption{The overhead, in terms of floats F (\textit{e.g.} trainable parameters) and binaries B (\textit{e.g.} binary values) \changemarker{in addition to the parameters of the neural network}. These values are an estimation, and are related to the additional values stored within the model once the training phase is over. For comparison, the number of floats parameters of a base network is indicated with the constant $b$, and the number of tasks is $M$.}
		\label{table:cl_memory}
		\resizebox{0.7\linewidth}{!}{%
			\begin{tabular}{|l|c|c|}
				\hline
				Method                          & F & B \\ \hline \hline
				Single Model                    &$(M - 1)\cdot b$&$0$\\ \hline
				SupSup                          &$0$&$M \cdot b$\\ \hline
				Pruning                         &$0$&$M\cdot b$\\ \hline
				Batch Ensemble                  &$ <0.1 \cdot b \cdot M$&$0$\\ \hline
				\nome{} (ours) &$0$&$< 0.001 \cdot b \cdot M$\\ \hline
			\end{tabular}%
		}
	\end{table}
	\begin{table}[t]
		\centering
		\caption{The number of bytes used by each method for two networks. The Naive approach is used to show the number of base memory required by each model, which is $b$. The additional memory required by each method depends on the number of tasks $M$. The results are correlated with the ones exposed in Table \ref{table:cl_memory}. Here, we set $F=64$, because we assumed that a float requires $64$ bites to be saved in the memory, and $B=1$, because each binary value requires $1$ bit. 
		}
		\label{table:cl_memory_example}
		\resizebox{0.7\linewidth}{!}{%
			\begin{tabular}{|l|c|c|} \hline
				Method & LeNet-5 & half VGG11  \\ \hline \hline
				Naive (b)              & $50550 \cdot 64$ & $2304864 \cdot 64$  \\ \hline 
				Single Model   &   $50550 \cdot 64 \cdot M$       &  $2304864 \cdot 64 \cdot M$          \\ \hline \hline
				
				SupSup & $50550 \cdot M$ & $2304864 \cdot M$  \\ \hline
				Pruning & $50550 \cdot M$ & $2304864 \cdot M$  \\ \hline
				Batch Ensemble  & $164 \cdot 64 \cdot M$   & $2499 \cdot 64 \cdot M$     \\ \hline
				
				\nome{}  (ours)        & $142\cdot M $   & $1376 \cdot M $  \\ \hline    
			\end{tabular}%
		}
	\end{table}
	Memory is a crucial aspect of any CL approach, because the number of tasks can grow over time, and so it is important to keep the required memory limited. 
	Table \ref{table:cl_memory} approximates the memory required by each method, \changemarker{by counting the number of additional values stored in the memory while taking into account the type of the number to save}. We partition memory into two components: the floats (F), and the binaries (B). \changemarker{The first one counts the additional values that must be stored as float values (e.g. the vectors in Batch Ensemble), while the second one counts only the values that can be saved as 0 or 1}. The split is necessary because a binary value requires less storage memory than a float (usually $\text{B}=64\text{F}$). 
	Between the methods, the worst are the one that apply binary masks over the whole networks (SupSup and Pruning), followed by BatchEnsemble and \nome{}, with the latter being capable of further reducing memory usage, making it negligible. 
	
	In Table \ref{table:cl_memory_example} we expose the approximated count of additional bits used by each method.
	We can see that our proposal is the best method, as said before. In fact, taking LeNet-5 as reference, we have that our method can store $355\ (\frac{50550}{142})$ masks, used to solve the corresponding number of tasks, using the same space that SupSup uses to store the mask for a single task. The second best method in terms of memory, BatchEnsemble, can store only $4$ tasks using the same space (due to the float values used by the method).
	
	In the end, the best methods are Batch Ensemble and our proposal, \nome{}. Both of the approaches requires small memory overhead (with the latter being better under this aspect). Taking in consideration both the accuracy results exposed before (Table \ref{table:classification}) and the memory Tables \ref{table:cl_memory} and \ref{table:cl_memory_example}, we conclude that our method is the best in terms of both memory requirements and accuracy obtained.    
	
	\subsubsection{\changemarker{Hyper-parameters of \nome{}}}
	\label{sec:exp_cl_ablation}
	In this section we study how the hyper-parameters of our method affect the results obtained on S-CIFAR10 dataset.

	\textbf{Local vs Global pruning}: in Table \ref{table:cl_ablation} the results of the various combinations are shown. The results show clearly that the local approach is always better than the global one. To understand why it happens we inspected the \changemarker{scaling vectors} produced by the method, and discovered that when using the global approach, the method tends to assign all the neurons in the first layers to the first task, because these are the ones which extract the features directly from the images.
	This prevents future tasks to adjust how these features are extracted from the images, and so they are forced to work with inappropriate high level features. On the contrary, using the local approach, a new task can always change the inactive neurons of the first layers, allowing custom high level features extraction. 
	   
	\textbf{Hard vs Soft extraction}: Table \ref{table:cl_ablation} also shown the difference between soft and hard extraction. In the local approach, we achieve better results using hard extraction approach, because using this schema the method is always capable of allocating new learnable neurons for the current task, while in the soft pruning case the number of active neurons may vary, and this could lead to a lack of neurons that can be used to solve properly the current task. This phenomenon is also visible when it comes to the global pruning, but the scores are not good enough in either case, due to the limitations exposed before. 
	
\begin{wrapfigure}{r}{7cm}
	\centering
	\includegraphics[width=6cm]{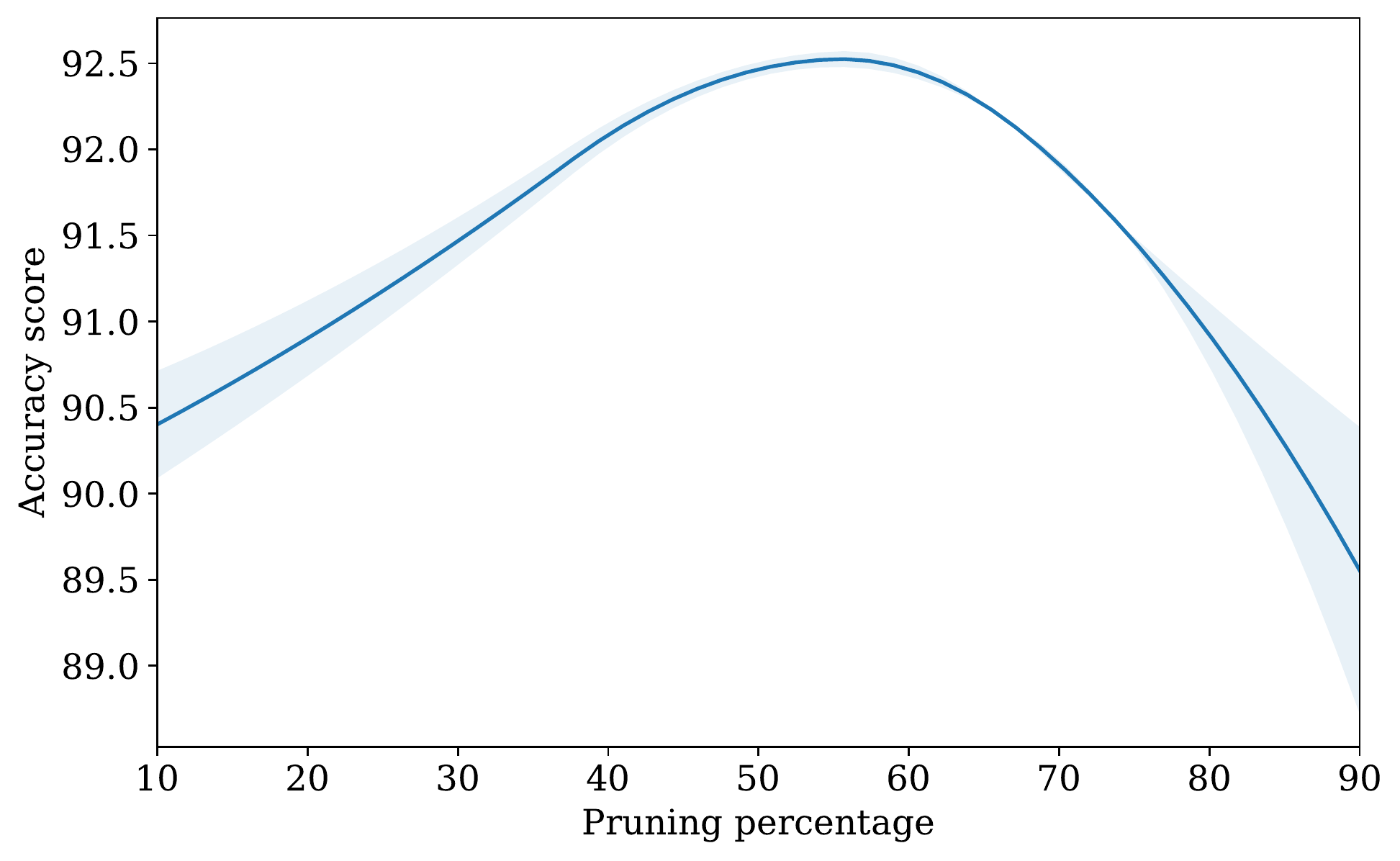}
	\caption{How pruning percentage affects the final score obtained on S-CIFAR10, with L-H setting.}
	\label{fig:cl_pruning}
	\end{wrapfigure}

	%
	\begin{table}[t]
		\centering
		\caption{The results obtained on S-CIFAR10 scenario by combining pruning techniques, which are: Hard (H) and Soft (S) extraction. Each one can be done locally in each layer (L) or globally (G). The results are averaged over two experiments, and best one is highlighted in bold.}
		\label{table:cl_ablation}
		\resizebox{0.3\linewidth}{!}{%
		\begin{tabular}{ll}
			\hline
			L-S&$92.14 \scalebox{.8}{\rpm 0.13}$ \\
			L-H& $\mathbf{92.50  \scalebox{.8}{\rpm 0.11}}$ \\ \hline
			G-S& $91.90 \scalebox{.8}{\rpm 0.31}$ \\
			G-H& $92.26 \scalebox{.8}{\rpm 0.22}$ \\
		\end{tabular}
		}
	\end{table}
	\textbf{Pruning percentage}: Figure \ref{fig:cl_pruning} shows how the accuracy score varies with the pruning percentage (the results are obtained using local-hard extraction). The pruning percentage must be set in such a way that the first task has all the computational capacity to be solved correctly. At the same time, this must be achieved using the smallest number of neurons, in order to leave some of them free for future tasks.  
	
	We can see that the worst results are obtained when the pruning percentage is too low, around $10\%$, or too high, around $90\%$, because, respectively, the space allocated for future tasks is small or the first one has a small amount of network to use (the first case is slightly preferable, because at least the first task can be solved efficiently, and this can also improves the accuracy on future tasks). 
	The best values are achieved with a percentage in $[50, 55]\%$, which is the final value used in all the experiments exposed before. 
	
	\section{Conclusion}
	
	In this paper we proposed a novel approach to build ensembles of deep models requiring less memory. Our approach, called \nome{}, is capable of extracting multiple networks from a single one, and use these to build a smaller ensemble model, which gives comparable results. Each network is different from the others, and contains the minimal set of structural connections from the original network that can be trained to achieve good accuracy while reducing the required memory. 
	
	We evaluated or proposal on a wide set of different experiments, including an evaluation of their robustness and diversity, showing it performs on-par or better than state-of-the-art approaches in the ensemble literature. 
	We also evaluated our method in a continual learning scenario, in which the model should be capable of solving a stream of tasks without losing the ability to solve past learned ones. Here too we compared our method to similar approaches, on a wide set of benchmarks, and we achieved the best scores while keeping the memory requirement negligible. 
	
	Future work will explore the time aspect of \nome{}, by focusing on techniques that can speed up the training process, while keeping the ability to extract multiple models from a single one, keeping the memory requirements contained.
    
    \bibliographystyle{elsarticle-num}
	\bibliography{main}

\end{document}